# ORANSight-2.0: Foundational LLMs for O-RAN

Pranshav Gajjar[1] and Vijay K. Shah[1,2]
[1]*NextG Wireless Lab*, North Carolina State University, and [2]WiSights Lab
{prgajjar and vijay.shah}@ncsu.edu, vijay.shah@wisightslab.com

***Abstract*—Despite the transformative impact of Large Language Models (LLMs) across critical domains such as healthcare, customer service, and business marketing, their integration into Open Radio Access Networks (O-RAN) remains limited. This gap is primarily due to the absence of domain-specific foundational models, with existing solutions often relying on general-purpose LLMs that fail to address the unique challenges and technical intricacies of O-RAN. To bridge this gap, we introduce ORANSight-2.0 (O-RAN Insights), a pioneering initiative to develop specialized foundational LLMs tailored for O-RAN. Built on 18 models spanning five open-source LLM frameworks — Mistral, Qwen, Llama, Phi, and Gemma — ORANSight-2.0 fine-tunes models ranging from 1B to 70B parameters, significantly reducing reliance on proprietary, closed-source models while enhancing performance in O-RAN-specific tasks. At the core of ORANSight-2.0 is RANSTRUCT, a novel Retrieval-Augmented Generation (RAG)-based instruction-tuning framework that employs two LLM agents — a Mistral-based Question Generator and a Qwen-based Answer Generator — to create high-quality instruction-tuning datasets. The generated dataset is then used to fine-tune the 18 pre-trained open-source LLMs via QLoRA. To evaluate ORANSight-2.0, we introduce srsRANBench, a novel benchmark designed for code generation and codebase understanding in the context of srsRAN, a widely used 5G O-RAN stack. Additionally, we leverage ORAN-Bench-13K, an existing benchmark for assessing O-RAN-specific knowledge. Our comprehensive evaluations demonstrate that ORANSight-2.0 models outperform general-purpose and closed-source models, such as ChatGPT-4o and Gemini, by *5.421%* on ORANBench and *18.465%* on srsRANBench, achieving superior performance while maintaining lower computational and energy costs. We also experiment with RAG-augmented variants of ORANSight-2.0 models and observe that RAG augmentation improves performance by an average of 6.35% across benchmarks, achieving the best overall cumulative score of 0.854, which is 12.37% better than the leading closed-source alternative. We thoroughly evaluate the energy characteristics of ORANSight-2.0, demonstrating its efficiency in training, inference, and inference with RAG augmentation, ensuring optimal performance while maintaining low computational and energy costs. Additionally, the best ORANSight-2.0 configuration is compared against the available telecom LLMs, where our proposed model outperformed them with an average improvement of 27.96%.**

***Index Terms*—O-RAN, ORANSight, 5G, LLM, ORANBench, srsRANBench, QLoRA, Foundational Models**

## I. INTRODUCTION

Open Radio Access Networks (O-RAN) have emerged as a transformative paradigm in the telecommunications industry, enabling greater flexibility, interoperability, and innovation by disaggregating hardware and software components [1]. However, the complexity of O-RAN architectures, coupled with the need for real-time automation, optimization, and troubleshooting, poses significant challenges. Large Language Models (LLMs) have shown immense potential in addressing these challenges by enabling intelligent decision-making, code generation, and natural language understanding [2] [3]. Despite this potential, the integration of LLMs into O-RAN remains nascent, primarily due to the lack of foundational LLMs tailored to the unique requirements of O-RAN systems.

***Related Works.*** Recent advancements in the telecom industry have seen growing adoption of LLMs [4]. They have been utilized for network modeling [5] and to obtain network configurations [6], for prediction tasks like traffic load prediction [7] and beamforming [8]. To analyze large-scale document corpora [9], and also as chatbot interfaces to facilitate development and help as an engineering assistant [10]. These works show the latent potential of such LLMs and how they can be incorporated into telecom, and we can assume the implications that these approaches have for O-RAN. A lot of the aforementioned work leverages prompt-tuning [5] or Retrieval-Augmented Generation [11] pipelines to instill domain knowledge in an LLM. There has also been work towards dataset curation and model training like Nikbakht et al. [12] that introduced TSpec-LLM, an open-source dataset aimed at enhancing LLM understanding of 3GPP specifications. Similarly, Maatouk et al. [9] proposed TeleLLMs, a series of specialized language models trained on telecommunications data, specifically 3GPP specifications, demonstrating improvements over general-purpose LLMs. Zou et al. [13] proposed TelecomGPT, a framework for building telecom-specific LLMs with tailored benchmarks and datasets. However, while these works contribute to LLM advancements in telecommunications, they do not include the O-RAN specifications or address any O-RAN-related challenges.

Recent efforts have been made to integrate LLMs into Open Radio Access Networks O-RAN and to bridge the domain gap between the pretraining of base LLMs and the specific requirements of O-RAN. In our previous work [11], we introduced ORAN-Bench-13K, a comprehensive benchmark designed to evaluate the performance of LLMs in O-RAN-specific tasks. This benchmark highlighted the limitations of existing models in handling O-RAN-related challenges. Additionally, we proposed a preliminary version of ORANSight, which addressed domain-based limitations by leveraging Retrieval-Augmented Generation (RAG) [11]. Similarly, Karbalaee Motalleb et al. [14] explored the use of LLMs to enhance the security of O-RAN architectures, emphasizing the necessity of domain-specific fine-tuning to mitigate vulnerabilities and threats. Furthermore, Lotfi et al. [15] demonstrated the effectiveness of LLM-augmented Deep Reinforcement Learning (DRL) for dynamic O-RAN network slicing, showcasing the potential of LLMs in improving decision-making processes within complex and evolving environments. NVIDIA also showcased an open-source O-RAN chatbot that follows a RAG-based approach with the NIM microservices and NVIDIA's LLM tools [16]. Despite these advancements, the existing approaches suffer from two significant drawbacks. First, none of these works have incorporated direct fine-tuning on O-RAN-specific text or addressed code generation for O-RAN in any capacity. Second, these approaches predominantly rely on closed-source proprietary models, which present several challenges. These models are plagued by privacy concerns, a lack of O-RAN knowledge, and the requirement for prompt and data sharing with the parent organization [9]. Additionally, they are entirely black-box models, offering no access to their training information, and cannot be deployed locally [17]. These limitations underscore the need for more transparent, domain-specific, and locally deployable solutions in the integration of LLMs within O-RAN.

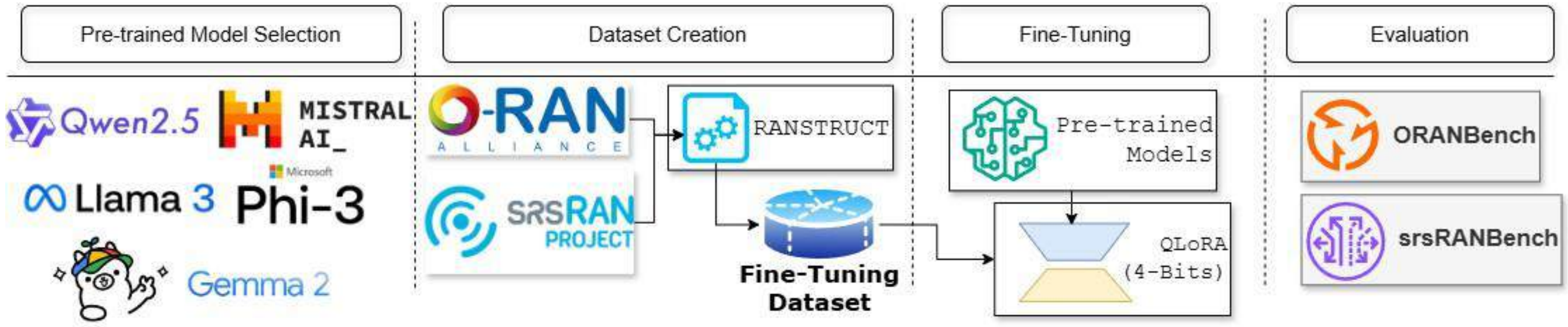

Fig. 1. Overall pipeline for ORANSight-2.0 models.

***Contributions.*** From the above discussion, it is evident that the integration of LLMs into O-RAN necessitates the development of domain-specific fine-tuning to address O-RAN's unique challenges, such as real-time network slicing, resource allocation, and fault diagnosis. While LLMs have demonstrated remarkable capabilities in general-purpose tasks, their application in O-RAN-specific scenarios is *constrained by the lack of open foundational models* trained on O-RAN-specific data. Furthermore, scaling LLMs for O-RAN applications presents additional challenges, including computational resource requirements, energy efficiency, and the need for high-quality, domain-specific datasets.

In this work, we make the following contributions.

- We present **ORANSight-2.0** [1], the first comprehensive effort to develop foundational LLMs tailored for O-RAN. Specifically, using the QLoRA (Quantized Low-Rank Adaptation) method [18], a parameter-efficient fine-tuning technique, we fine-tune 18 LLM models from five leading open-source LLM model families, namely, Mistral, Qwen, Llama, Phi and Gemma - ranging from **1B** to **70B** parameters. ORANSight-2.0 addresses the limitations of closed-source models and provides open-source alternatives optimized for O-RAN tasks. We not only work towards foundational models that are proficient for the specifications, but we also emphasize the code generation and understanding capabilities.
- We introduce **RANSTRUCT**, a Retrieval-Augmented Generation [11] (RAG)-based instruction-tuning framework that leverages two LLM agents to generate high-quality, O-RAN-specific instruction-tuning datasets.
- Finally, we present **srsRANBench**, a novel benchmark designed for code generation and codebase understanding for the srsRAN project [19], an open-source, widely used 5G O-RAN stack not only used by academia for R&D purposes but also by industry for 5G O-RAN network deployments across the US and Europe. srsRANBench enables the evaluation of various LLMs in 5G O-RAN-specific coding tasks.

***Experiment.*** We extensively evaluate the performance of ORANSight-2.0 against state-of-the-art closed-source models, including ChatGPT-4o [20], and Gemini 1.5 [21]. We thoroughly evaluate the efficacy of RANSTRUCT and the impact of RAG-based ORANSight augmentation. The energy characteristics associated with deploying ORANSight or performing further fine-tuning are also analyzed. Additionally, we evaluate our model's performance against various telecom LLM baselines. Furthermore, all ORANSight-2.0 models [2] and dataset benchmarks [3] are publicly available, fostering reproducibility and enabling future research in O-RAN and AI applications.

***Paper Organization.*** The remainder of this paper is organized as follows: Section II explains the background and motivation behind this work, and Section III details ORANSight-2.0, including the design of the RANSTRUCT framework. Section IV introduces dataset benchmarks, including srsRANBench, and describes the evaluation methodology. Section V presents the experimental results, including comparisons with closed-source models and energy consumption analysis. Section VI highlights the limitations of this work, and Section VII concludes the paper and discusses future research directions.

[1]In the rest of the paper, we will use ORANSight-2.0 and ORANSight interchangeably to refer to the proposed foundational LLMs for O-RAN.

[2]All 18 models along with the required code for inference available in Hugging Face, Link: https://huggingface.co/NextGLab

[3]The ORANBench13K is available at: https://github.com/prnshv/ORAN-Bench-13K, and the srsRANBench is available at: https://github.com/prnshv/srsRANBench.

## II. Background and Motivation

Large Language Models (LLMs) have demonstrated remarkable capabilities in natural language understanding and generation, making them a promising solution to address various challenges within Open Radio Access Networks (O-RAN). However, the application of LLMs in O-RAN is still in its early stages, primarily due to the absence of foundational models tailored to the unique requirements of O-RAN systems. LLMs are typically trained using a next-token prediction objective, where they learn to predict the next word or subword in a sequence based on vast and diverse datasets spanning fields such as medicine, history, and more [9]. This training process enables the models to understand the statistical relationships between tokens, which are the smallest units of text, and develop a broad, general-purpose understanding of language. However, as highlighted in [22], when an LLM is intended for use in a specialized domain, adapting the model to that domain becomes imperative. This is because the patterns and distributions of tokens in domain-specific texts often differ significantly from those in the general-purpose datasets the model was originally trained on [22]. A prominent way to achieve this adaptation is by training the model with high-quality domain-specific data, which aligns the model with the unique token distribution of the target domain, leading to improved performance in downstream tasks.

### A. Challenges of Current LLM Solutions for O-RAN

Currently, LLM solutions for O-RAN are predominantly built on Retrieval-Augmented Generation (RAG) [11] [16], which is a primitive approach to instilling domain knowledge in LLMs. RAG relies on querying external knowledge bases for every prompt, which can provide contextually relevant responses but is inefficient for real-time O-RAN applications due to its computational overhead and dependency on constant retrieval. In contrast, fine-tuning has been extensively documented to enable a model to internalize domain-specific knowledge, reducing the need for external queries and enabling faster, more accurate responses [9] [23]. This makes fine-tuning a more suitable approach for O-RAN applications, where real-time performance and efficiency are critical.

### B. Absence of Domain-Specific Datasets for O-RAN

A significant challenge in adapting LLMs to O-RAN is the lack of datasets specifically designed for fine-tuning LLMs in this domain. While datasets like TSpec-LLM [12] and TelecomGPT [13] exist for broader telecom standards, they do not address the specific requirements of O-RAN. This absence of an O-RAN-specific dataset necessitates the creation of an approach that would allow for the generation of a fine-tuning set tailored to our needs. Furthermore, to the best of our knowledge, no existing work addresses the coding abilities of these models in O-RAN, highlighting the importance of developing a dataset that also addresses this limitation. It is also important to note that manual annotation and data generation are extremely time-consuming and expensive, as evidenced by the papers [24] [25], and developing an approach that can generate quality O-RAN-specific datasets without any human intervention is a crucial task.

### C. Efficiency of Fine-Tuning Pre-Trained Models

Fine-tuning pre-trained models is a more efficient approach than training LLMs from scratch for O-RAN-specific tasks. Pre-trained models already possess a broad understanding of language, and fine-tuning allows them to specialize in O-RAN without the need for extensive computational resources. This trend is evident in multiple studies that emphasize fine-tuning as a superior alternative to training from scratch [26]. Additionally, several LLMs have been pre-trained to follow instructions, achieving superior results in various downstream tasks [9] [23]. This further underscores the effectiveness of fine-tuning as a strategy for adapting LLMs to specialized domains like O-RAN.

### D. Parameter-Efficient Fine-Tuning (PEFT) Techniques

Parameter-efficient fine-tuning (PEFT) techniques are essential for optimizing the fine-tuning process, particularly in resource-constrained environments. PEFT methods focus on updating only a small subset of the model's parameters, significantly reducing computational overhead while maintaining performance [23]. A complementary approach to further optimize efficiency without significant performance degradation is reducing precision. Model weights are typically stored in full precision (32-bit floating point), but lower-precision formats, such as 4-bit quantization, drastically reduce memory and computational costs [27]. We draw inspiration from the work of [18], which proposed Quantized Low-Rank Adaptation (QLoRA) and demonstrated a PEFT paradigm with performance comparable to full-precision networks but with substantially lower memory requirements.

Taking these considerations into account, we develop ORANSight-2.0 by first curating datasets specifically tailored to the O-RAN ecosystem. We then employ advanced fine-tuning techniques to build foundational models optimized for O-RAN-related text and code generation. This targeted approach bridges the gap between general-purpose LLMs and the unique, specialized demands of the O-RAN domain, offering more effective and efficient solutions for this critical O-RAN domain use case.

## III. ORANSight-2.0: Foundational O-RAN LLMs

This section details the methodology employed in developing ORANSight-2.0 – (i) the selection of 18 LLM models across five leading, open-source LLM model families, (ii) the RANSTRUCT framework, and (iii) the QLoRA-based fine-tuning of LLMs. Figure 1 shows the overall pipeline for ORANSight-2.0 models.

### A. Pre-trained Model Selection

LLM models are generally released as a family across different parameter sizes while maintaining the same training methodology. Existing work in telecom typically targets a single model family or smaller LLMs to achieve a downstream task [28] or focuses on smaller LLMs ($< 10$B parameters) for an initial study [9]. It is important to note that since each publicly available model family undergoes different *pretraining*, a comprehensive evaluation should include as many models as possible. A model's overall capability is also strongly correlated with its *parameter size*, as deeper models typically exhibit superior representational abilities compared to smaller ones. This trend is seen in all the model releases, as the larger models with the same training method outperform their smaller counterparts [29] [30] [31]. While studying smaller LLMs can provide insights into their applicability, developing a highly impactful foundational model requires targeting larger LLMs.

Another key aspect of LLM performance is its *context length*, which determines how much prior information, or the maximum number of tokens, a model can effectively utilize in a single forward pass. Larger models often feature extended context windows, allowing them to better capture dependencies across longer sequences, which is crucial for tasks involving reasoning, document comprehension, and telecom-related applications like log analysis [32]. Recent advancements have significantly expanded the context lengths of modern LLMs, with some models now supporting up to 128K tokens. However, the ability to retain and recall relevant information across long contexts varies among models, making it a crucial factor to include larger models for ORANSight-2.0. Therefore, we select **18** state-of-the-art models covering a broad range of parameter sizes **(from 1B to 70B)** that can be effectively fine-tuned using our computational resources. All pre-trained LLMs serving as the foundation for ORANSight-2.0 have demonstrated exceptional performance across diverse general text understanding and code generation benchmarks. The chosen models, along with their parent organizations, parameter sizes, and context lengths, are listed in table I.

The chosen model families have been well-received in the literature for general LLM tasks, but they also showcase relevance and early adoption in the telecom and LLM domains. With the aforementioned related works from Section I, Tele-LLM [9] focused on Phi, Gemma, and the LLaMA family; TelecomGPT [13] utilized LLaMA and Mistral, and the NVIDIA-O-RAN chatbot [16] also focused on the LLaMA and Mistral models.

### B. RANSTRUCT

ORANSight-2.0 addresses the problem of creating an instruction-based fine-tuning set by RAG-based INSTRUCTion Generation, dubbed, **RANSTRUCT**. The proposed framework is shown in fig 2, and at its core, we leverage two LLM agents, one for generating questions or instructions, and one LLM for the corresponding answers or responses. The framework is built on the fundamentals of RAG, which guarantees an accurate response if the scope of the instruction is within the RAG database [35] [36].

Implementing RAG requires an efficient mechanism for storing and retrieving relevant information. Since large documents cannot be processed in their entirety, they are first segmented into semantically meaningful chunks to facilitate precise retrieval. These chunks ensure that when a query is issued, only the most relevant portions of the dataset are accessed, reducing extraneous information and improving retrieval efficiency [11].

To enable effective search and retrieval, each chunk is converted into a dense vector representation using a pre-trained embedding model, which has an associated context length or the maximum number of tokens that it has been trained to represent as embeddings. This transformation captures the semantic properties of the text, allowing for similarity-based retrieval based on contextual relevance [11]. The embeddings are then indexed in the RAG database, which supports high-speed similarity searches across large datasets. In our instruction generation process, questions are formulated such that they are explicitly answerable based on the retrieved chunks. Similarly, during response generation, the model utilizes retrieved-context from the RAG database to produce outputs that are both accurate and contextually grounded [11].

For ORANSight-2.0, we select the `Mistral-7B-Instruct-0.3` and the `Qwen-2.5-Instruct-1.5B` models due to their efficiency and compatibility with RAG pipelines. Qwen, in particular, has been extensively documented to work effectively with RAG [35], enabling faster outputs and efficient dataset generation despite its smaller size. This choice ensures that the framework operates efficiently while maintaining high-quality outputs. We run the framework for two instances, with the O-RAN specifications and individual files from the srsRAN project. Below, we outline the workflow of RANSTRUCT, as depicted in Figure 2:

1. Input Processing. The RANSTRUCT framework begins with two primary O-RAN data sources: **O-RAN Specifications** and the **srsRAN Code Files**. The O-RAN Specifications consist of 116 documents (a combination of PDFs and Word files), averaging 21,778 words per document, totaling 2.53 million words. These specifications form the foundation for generating structured instruction-tuning datasets. The srsRAN Code files, on the other hand, comprise C++ source files from the srsRAN project, a widely used open-source implementation of the RAN stack by academia and industry. This codebase includes 4,980 files and 4.68 million words.

To prepare the data, we employ a **Recursive Splitter**[4] to generate semantically meaningful chunks. For the O-RAN Specifications, two types of chunks are created: **RAG Chunks** (1,024 tokens) for efficient document retrieval within the RAG pipeline and **LTG (Long-Text Generation) Chunks** (4,096 tokens) for generating high-quality question-answer pairs. The specific token values are chosen to comply with the embedding generator's requirements. In contrast, for the srsRAN codebase files, entire codebase files are used instead of chunking to preserve semantic context.

[4]A Recursive Splitter refers to a technique that breaks down complex tasks or large datasets into smaller, more manageable parts, processed iteratively. This helps in improving the efficiency and performance of LLMs by focusing on sub-problems, especially when dealing with hierarchical structures or large volumes of data.

TABLE I
COMPARISON OF VARIOUS LARGE LANGUAGE MODELS AND THEIR SPECIFICATIONS.

| Model | Developer | Parameter Sizes | Pretraining Details | Context Length |
|---|---|---|---|---|
| Qwen 2.5 [29] | Alibaba | 1.5B, 3B, 7B, 14B, 32B | Trained on an 18-trillion-token dataset with supervised fine-tuning using over 1 million samples and multistage reinforcement learning to enhance instruction following and reasoning abilities. | Up to 128K tokens; generates up to 8K tokens |
| Gemma 2 [30] | Google | 2B, 9B, 27B | Utilized datasets ranging from 2 trillion to 13 trillion tokens, including web documents, code, and scientific articles. Employed knowledge distillation and architectural modifications like interleaving local-global attention and group-query attention. | 8K tokens |
| Mistral [33] | Mistral AI | 7B, 12B (Nemo), 22B (Small), 8x7B | The company has pioneered Grouped-Query Attention (GQA) and Sliding Window Attention (SWA) for efficient performance in Mistral models. Despite being open source, specific pretraining dataset details are not publicly available. | 7B, 22B, 8x7B: 32K tokens; 12B (Nemo): 128K tokens |
| Phi 3 [31] | Microsoft | 3.5: Mini; 3: Medium | These models are trained on a 3.3-trillion-token dataset comprising heavily filtered web data, curated educational content, and synthetic "textbook-like" data to enhance language understanding and reasoning capabilities. | 4K tokens |
| Llama 3 [34] | Meta AI | 3.1: 8B, 70B; 3.2: 1B, 3B | Leveraged a dataset comprising 15.6 trillion tokens, focusing on diverse language tasks to enhance general language understanding and generation capabilities. | 3.1: 128K tokens; 3.2: 8K tokens |

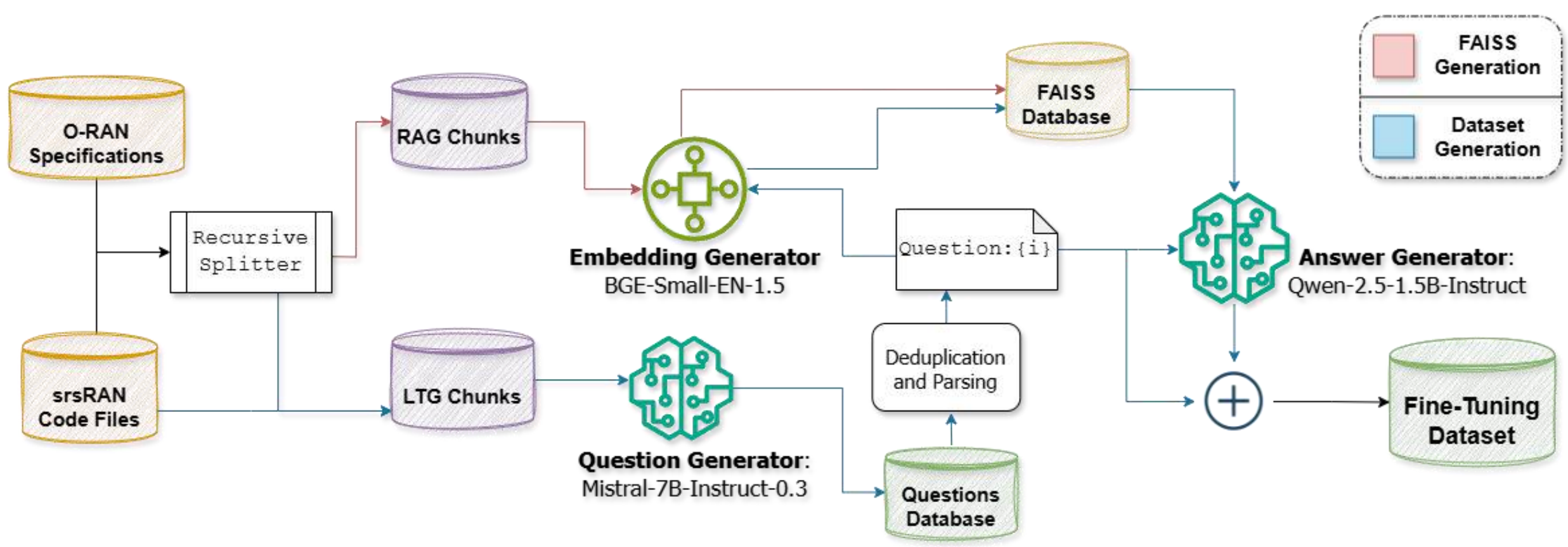

Fig. 2. The proposed RANSTRUCT Framework

2. FAISS Construction. As RAG works through a database of embeddings we encode the RAG chunks using the `BGE-Small-EN-1.5` embedding model [37] to create dense vector representations, that are then indexed using a **FAISS** (Facebook AI Similarity Search) database. We selected this particular combination due to its documented performance in O-RAN domain from our previous work [11] and its alignment with the open-source nature of our work. The FAISS Construction concludes when we have exhausted all the RAG chunks and we process with the next step.

3. Question Generation. The third stage of RANSTRUCT involves generating a structured database of questions that are crucial for the knowledge base of an O-RAN LLM. For each LTG chunk, we prompt the question generator model (`Mistral-7B-Instruct-0.3`) to generate multiple unique questions that are strictly within the scope of the LTG chunk and can be accurately answered if the same information is provided as context. The model is prompted to generate the questions as a list which can be parsed by a separate script.

Once we have a list of questions across all possible LTG chunks, we subject this database to de-duplication and parsing. De-duplication is achieved by transforming the list of questions into a dictionary, where the questions serve as keys. This approach inherently removes duplicates, as dictionaries cannot have duplicate keys, ensuring that unique questions are retained. Parsing is achieved by filtering out truncated, malformed, or invalid questions to maintain dataset integrity.

This process ensures that only high-quality, unique questions are retained for the final instruction-tuning dataset.

4. Answer Generation. Once we have unique and high-quality questions, we leverage the `Qwen-2.5-Instruct-1.5B` to create the corresponding response. Each question is subjected to the embedding generator to obtain its semantic representation. We leverage the embeddings through the FAISS database to obtain the top 3 chunks. The Qwen model is then provided with the question and the retrieved documents as context to generate a response. Since the questions are generated while strictly adhering to the scope of an LTG chunk, the retrieved documents inherently contain the necessary information to provide accurate answers. This ensures that the context provided by the Qwen model is always relevant and sufficient for generating high-quality responses. Finally, the generated answer and the original question are appended to the fine-tuning set.

**Dataset Metrics:** After exhausting all the questions, we obtain the dataset for training ORANSight-2.0 models which contain **1,51,500** unique instruction and answer pairs across the srsRAN project codebase and the O-RAN specifications, leading to a total of **29,864,388** words. Out of this, **88,766** samples or 58.591% are related to the srsRAN project codebase, and **62,734** or 41.409% are related to the O-RAN Specifications. Throughout the rest of the paper, this dataset is referred to as the RANSTRUCT dataset.

### C. QLoRA

Fine-tuning of each selected LLM model on the RANSTRUCT dataset in ORANSight-2.0 is performed through the QLoRA (Quantized Low-Rank Adaptation) technique. As introduced in section II, it is a PEFT method that enables the training of large language models in resource-sensitive scenarios without compromising performance. At its core, QLoRA combines two key techniques: *Quantization* and *Low-rank adaptation (LoRA)* [18]. Quantization reduces the memory footprint by mapping the original 16-bit floating-point weights to 4-bit integers, significantly lowering computational and storage requirements [18]. Low-rank adaptation, on the other hand, introduces trainable low-rank matrices to adapt the pre-trained model weights, drastically reducing the number of trainable parameters. Mathematically, for a pre-trained weight matrix $W \in \mathbb{R}^{m\times n}$, LoRA introduces a low-rank decomposition where the weight update is represented as $\Delta W = AB$, with $A \in \mathbb{R}^{m\times r}$ and $B \in \mathbb{R}^{r\times n}$, where $r \ll \min(m, n)$ [38]. Instead of modifying $W$ directly, the adapted transformation is expressed as:

$$Y = X(W + sAB), \tag{1}$$

where $s$ is a scaling factor. This decomposition ensures that only a small number of parameters are updated during fine-tuning, preserving the efficiency of the model while maintaining expressiveness. QLoRA further enhances this efficiency by proposing 4-bit quantization (represented as NormalFloat or NF4) to store pre-trained model weights, which are dequantized to 16-bit (represented as bfloat16 or BF16) during computation [18]. The dequantization process is denoted as $dD$, where the stored NF4 weights are first scaled using a 32-bit constant (denoted as $s^1_{FP\,32}$and then using an 8-bit constant (denoted as $s^2_{4\text{-bit}}$), this double step process greatly reduces memory consumption, allowing us to fine-tune larger ORANSight-2.0 models on our computational hardware. Finally, the forward pass is performed as

$$Y_{BF\,16} = X_{BF\,16} \cdot \mathrm{dD}(s^1_{FP\,32}, s^2_{8\text{-bit}}, W_{NF\,4}) + (XAB)_{BF\,16} \tag{2}$$

where $X_{BF\,16}$ represents the input in BF16 precision, $W_{NF\,4}$ are the stored 4-bit weights, and $X_{BF\,16}A_{BF\,16}B_{BF\,16}$ is analogous to the the LoRA-based low-rank update.

## IV. Experimental Setup

The hardware configuration we leverage consists of an *Intel(R) Core(TM) i9-14900KF* CPU with 62 GB of RAM, paired with an *NVIDIA GeForce RTX 4090 GPU* with 24 GB of GDDR6X memory. All ORANSight-2.0 models are trained with the **unsloth** library [39], and a 4-bit precision except for Mistral 8x7B and Llama 70B, which are trained using the **AQLM** library [40] and a 2-bit precision. We rely on AQLM (Adapter-based Quantized Language Models) [5] due to additive quantization and the possibility of working with larger models in the same setup.

As shown in sections II and III, we primarily leverage a 4-bit precision as it has been documented throughout the literature to be the best tradeoff for generation fidelity and compute resources. It allows us to pursue a thorough comparative analysis that is shown in section V. We train our models for a single epoch with the maximum possible batch size of up to 8, which is determined by the available GPU memory. The rank and alpha hyperparameters for the QLoRA configuration are model-specific, with the maximum values set to 256 for both. These values are chosen based on the model architecture and the memory available for the best possible performance.

To further optimize resource utilization, we employ data packing during training. *Packing involves concatenating multiple shorter sequences into a single longer sequence, thereby reducing the number of padding tokens required*. This technique helps us maximize GPU utilization by ensuring that the computational resources are fully utilized for meaningful computations rather than processing padding tokens [41].

### A. Benchmarks

To evaluate the performance of the ORANSight-2.0 models, we leverage and extend our previous work, ORAN-Bench-13K [11], and introduce a new benchmark, srsRANBench, designed specifically for code generation and codebase understanding for the srsRAN Project. Both benchmarks are carefully curated to ensure a reliable and thorough evaluation while maintaining computational efficiency.

[5]AQLM is a method that combines adapter layers and quantization to efficiently fine-tune large language models with fewer resources, reducing both memory and computational requirements.

1. ORANBench We leverage ORAN-Bench-13K [11], our previously developed benchmark, consisting of 13,952 meticulously curated multiple-choice questions generated from the O-RAN specification documents across three difficulty levels: *easy*, *intermediate*, and *difficult*. For this work, we create ORANBench by randomly selecting a subset of 500 samples from each difficulty category, resulting in a balanced evaluation set of 1,500 questions. This streamlined version retains the diversity and rigor of the original benchmark while reducing computational overhead, enabling efficient evaluation.

2. srsRANBench Building on the framework used for ORAN-Bench-13K, we introduce srsRANBench, a novel benchmark focused on code generation and codebase understanding for the srsRAN project. Using the same methodology, we curate a dataset of 1,502 samples by randomly selecting cpp files from the entire srsRAN codebase. This benchmark evaluates the ability of LLMs to generate syntactically and semantically correct code, as well as their understanding of the codebase. The random selection process ensures a representative and unbiased evaluation, while the manageable size of the dataset allows for computationally efficient testing.

Both Benchmark samples are showcased in Appendix A for ORANBench and Appendix B for srsRANBench.

### B. RAG-based ORANSight-2.0 augmentation for inference

As RAG is usually perceived as a tool to augment the outputs of an LLM with domain knowledge, it can also be used on top of a fine-tuned LLM to enhance its performance. RAG adds relevant context to the model inputs, which can help the model figure out the correct response by grounding its answers in retrieved information, reducing hallucinations, and improving accuracy, especially for knowledge-intensive tasks [35]. We adapt RAG for ORANSight-2.0 inference from the first version of ORANSight as described in our previous work [11], further advancing the state-of-the-art in O-RAN-specific tasks. The Inference RAG pipeline leverages a FAISS database constructed using the open source `bge-large-en-v1.5` embedding model [37], which supports the O-RAN Specification and the srsRAN codebase files, totaling **88,808 chunks** and **7,236,372 words**.

### C. Energy Consumption

We calculate the energy characteristics for table III by leveraging the `Codecarbon` package [42]. We use the package to observe the energy usage across training, standard inference, and inference with RAG. Codecarbon allows us to monitor detailed energy characteristics across the graphics processing unit (GPU), Central Processing Unit (CPU), and Random Access Memory (RAM). This enables a granular comparison between different model operations, providing insights into the trade-offs between performance improvements from RAG and the associated energy costs. The energy metrics also enable comparing ORANSight-2.0 models in computational efficiency, supporting informed decision-making in model selection for real-world deployment.

### D. RANSTRUCT Efficacy

We also thoroughly evaluate the effect of finetuning with our proposed RANSTRUCT framework by evaluating the base LLMs without any finetuning with an identical experimental setup. All the base-LLMs are available in Instruction tuned 4-bit precision from the unsloth [39] model hub, allowing a fair comparative analysis.

### E. SOTA Telecom LLMs

As shown in Section I, there are a handful of works that have targeted telecom and language models, and we also include a thorough comparison of ORANSight-2.0 against the available telecom LLMs. The tested models include TSPec-LLM [12][6], Tele-LLM [9][7], TSpec-LLM [12][8], and the NVIDIA-ORAN chatbot [16][9]. We use the best representative variant for each of these telecom LLMs, and we include a holistic comparison with all publicly available models. The only exception is TelecomGPT [13] where no open source or replicable implementation is available.

## V. Results

From the table II, we can observe that from the 18 ORANSight-2.0 models, 12 models outperform Gemini 1.5 Flash 8B, 10 models outperform Gemini 1.5, and 11 models outperform ChatGPT-4o Mini. However, we primarily aim to compare our work with ChatGPT-4o (4o) which is widely regarded as the best LLM model for every use case and can be assumed to be the deepest model in our comparative analysis. A total of 6 models outperform 4o for the cumulative score, and 11 models outperform 4o on the srsRANBench. Across the 6 individual models, ORANSight-Qwen-2.5 32B, achieves the highest cumulative score of 0.794, surpassing 4o (0.760) by 4.47%, it excels in both O-RAN-specific tasks and coding challenges, scoring 0.793 on ORANBench and 0.796 on srsRANBench. Following closely, ORANSight-Gemma-2 27B attains a cumulative score of 0.829, exceeding 4o by 9.08% with exceptional performance on srsRANBench (0.911) and a comparable result on ORANBench (0.747). Two Mistral Models, ORANSight-Mistral 22B (Small) and ORANSight-Mistral 12B (Nemo) outperform 4o with cumulative scores of 0.790 and 0.772, respectively. The smallest models to outperform the baseline are ORANSight-Gemma-2 9B and ORANSight-Qwen-2.5 7B, showcasing the best performance with substantially fewer parameters.

For the srsRANBench, the ORANSight-Gemma-2 27B model achieves the highest score (0.911) which is 12.7%

[6] The original TSpec-LLM paper leverages `Gemini 1.0 Pro`, now retired; per the official migration notice at https://cloud.google.com/vertex-ai/generative-ai/docs/learn/model-versions, we upgrade to `gemini-2.0-flash` for all experiments.

[7] For Tele-LLM we use the `Llama-3-8B-Tele-it`, which is reported as the strongest variant in their open-source release.

[8] The original TSpec-LLM paper fine-tunes `Gemini 1.0 Pro`, now retired; per the official migration notice at https://cloud.google.com/vertex-ai/generative-ai/docs/learn/model-versions

[9] We replicate the implementation from their Github: https://github.com/NVIDIA/GenerativeAIExamples/tree/main/community/oran-chatbot-multimodal

TABLE II
ORANSIGHT-2.0 AGAINST TWO CLOSED-SOURCE LLM BASELINES ACROSS VARIOUS MODEL SIZES

| **Model Family** | **Parameters (B)** | **ORANBench** | | | | **srsRANBench** | **Cumulative Score** |
|---|---|---|---|---|---|---|---|
| | | Easy | Medium | Difficult | Average | | |
| ORANSight-Gemma-2 | 2 | 0.698 | 0.618 | 0.584 | 0.633 | **0.848** | 0.74 |
| | **9** | 0.822 | 0.718 | 0.680 | **0.740** | **0.834** | **0.787** |
| | **27** | 0.818 | 0.722 | 0.702 | **0.747** | **0.911** | **0.829** |
| ORANSight-Mistral | 7 (v3) | 0.730 | 0.618 | 0.612 | 0.650 | **0.863** | 0.756 |
| | **12 (Nemo)** | 0.784 | 0.710 | 0.650 | 0.715 | **0.828** | **0.772** |
| | **22 (Small)** | 0.802 | 0.700 | 0.658 | 0.720 | **0.860** | **0.79** |
| | 8x7 (AQLM) | 0.780 | 0.654 | 0.626 | 0.687 | 0.753 | 0.72 |
| ORANSight-Llama | 3.2: 1 | 0.438 | 0.350 | 0.364 | 0.384 | 0.565 | 0.474 |
| | 3.2: 3 | 0.708 | 0.582 | 0.540 | 0.610 | 0.768 | 0.689 |
| | 3.1: 8 | 0.728 | 0.730 | 0.618 | 0.692 | 0.728 | 0.71 |
| | 3.1: 70 (AQLM) | 0.820 | 0.680 | 0.652 | 0.717 | 0.761 | 0.739 |
| ORANSight-Phi | 3.5: Mini | 0.716 | 0.670 | 0.658 | 0.681 | 0.754 | 0.718 |
| | 3: Medium | 0.754 | 0.710 | 0.682 | 0.715 | **0.783** | 0.749 |
| ORANSight-Qwen-2.5 | 1.5 | 0.660 | 0.586 | 0.606 | 0.617 | **0.804** | 0.71 |
| | 3 | 0.704 | 0.642 | 0.628 | 0.658 | **0.857** | 0.758 |
| | **7** | 0.788 | 0.720 | 0.696 | 0.735 | **0.837** | **0.786** |
| | 14 | 0.804 | 0.730 | 0.684 | 0.739 | 0.761 | 0.75 |
| | **32** | 0.856 | 0.784 | 0.738 | **0.793** | **0.796** | **0.794** |
| (Baseline) ChatGPT | 4o-mini | 0.766 | 0.727 | 0.677 | 0.723 | 0.755 | 0.739 |
| | 4o | 0.792 | 0.760 | 0.693 | 0.752 | 0.769 | 0.76 |
| (Baseline) Gemini | 1.5:8B | 0.723 | 0.665 | 0.631 | 0.673 | 0.767 | 0.72 |
| | 1.5 | 0.743 | 0.707 | 0.669 | 0.706 | 0.775 | 0.741 |

better than the average model and outperforms 4o (0.769) by 18.5%. The overall best model ORANSight-Qwen-2.5 32B also performs well, with a score of 0.796, which is lesser than ORANSight-Gemma-2 27B but is still better than our baseline. For ORANBench, the ORANSight-Qwen-2.5 32B leads with an average score of 0.793, followed by ORANSight-Gemma-2 27B (0.747) and ORANSight-Mistral 22B (Small) (0.720). These models consistently outperform closed-source counterparts, particularly in the difficult category, where ORANSight-Qwen-2.5 32B achieves a score of 0.738. Ideally, we expected consistency in performance across the two benchmarks, but the pertaining of the base LLM can play a role in the post-fine-tuning result, and we can assume that the proficiency of ORANSight-Gemma-2 27B in general coding tasks translated well for srsRAN.

Performance generally increases linearly with parameter size, with larger models like ORANSight-Qwen-2.5 32B and ORANSight-Gemma-2 27B achieving the highest scores. However, diminishing returns appear beyond a certain threshold, as seen with the ORANSight-Mistral 8x7B (AQLM) model, which underperforms despite its larger size. The reason for diminishing performance after 32B can be attributed to heavy quantization as the AQLM models were trained using 2bits instead of a 4bit setup. This was employed due to the limited computing resources, but the AQLM models gave a comparable performance to 4o-mini and Gemini 1.5: 8B. Smaller models, such as ORANSight-Qwen-2.5 3B, demonstrate that fine-tuning can significantly enhance performance even with limited parameters, making them viable for resource-sensitive scenarios.

To compare the performance across the different model families, ORANSight-Qwen-2.5 emerges as the top performer, with all variants (1.5B to 32B) surpassing ChatGPT-4o Mini and consistent performances for the cumulative scores. ORANSight-Gemma-2 particularly outperforms every model for srsRAN and two out of three models also provide superior performance on ORANBench. For the Mistral series, the 22B (Small) model, outperforms 4o Mini and competes closely with 4o, though the 8x7B (AQLM) model struggles to scale effectively. The ORANSight-LLaMA series delivers mixed results, with the 70B (AQLM) model performing well, while the smaller variants (1B and 3B) struggle with ORANBench, emphasizing the importance of model size. Lastly, the Phi family performs similarly to LLaMA, with both models usually underperforming across ORANBench and srsRANBench when compared against the other ORANSight-2.0 models. To further evaluate the performance of ORANSight-2.0 with RAG-based augmentation and for energy consumption, we select four models across different parameter ranges that gave the best cumulative performance:

1) **0-10B:** ORANSight-Gemma-2 9B
2) **11-20B:** ORANSight-Mistral-Nemo
3) **21-30B:** ORANSight-Gemma-2 27B
4) **31-40B:** ORANSight-Qwen-2.5 32B

### A. *RAG for Inference*

The performance of ORANSight-2.0 with RAG-based augmentation, as shown in Figure 3, highlights that RAG enhances the performance of fine-tuned LLMs and advances the state-of-the-art for both ORANBench and srsRANBench. ORANSight-Mistral-Nemo (12B) shows a 6.35% improvement in the cumulative score (0.821 with RAG vs. 0.772 without), with a 20.3% gain in the Difficult category of ORANBench. ORANSight-Gemma-2 9B achieves a 5.97% increase in cumulative score (0.834 with RAG vs. 0.787 without), with consistent improvements across all categories, particularly in srsRANBench (4.8% increase). However, ORANSight-Gemma-2 27B experiences a slight 1.08% drop in the cumulative score (0.820 with RAG vs. 0.829 without), likely due to low LLM recall, where the retrieval mechanism may not align well with the model's internal knowledge. ORANSight-Qwen-2.5 32B exhibits the most significant improvement, with a 7.42% increase in the cumulative score (0.854 with RAG vs. 0.795 without) and a 19.0% gain in the Difficult category of ORANBench, despite a minor drop in srsRANBench, suggesting a trade-off between O-RAN knowledge and coding tasks.

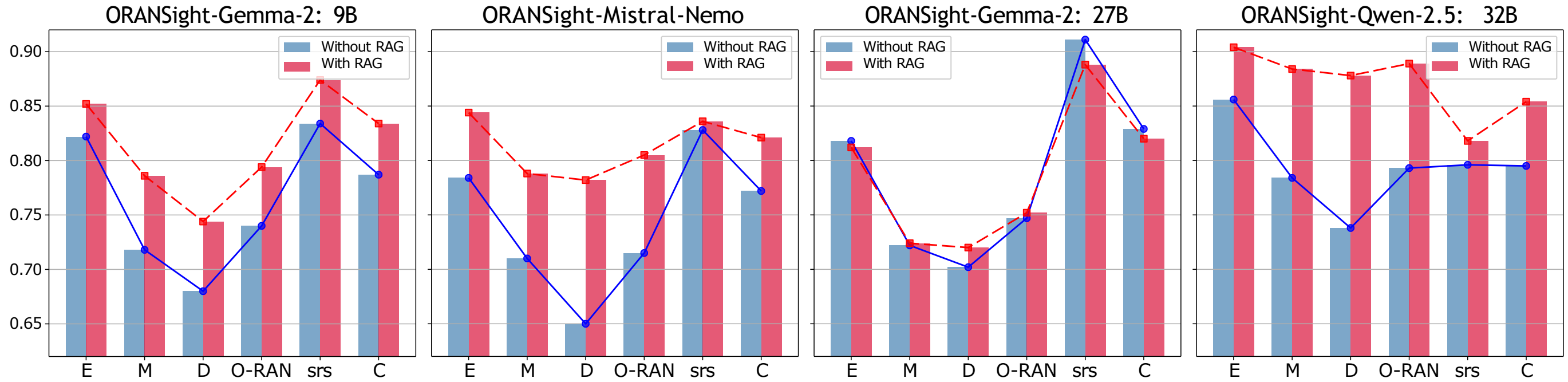

Fig. 3. Performance comparison of different model families with and without RAG. The categories on the x-axis represent different evaluation criteria: 'E', 'M', and 'D' correspond to Easy, Medium, and Difficult samples from ORANBench. 'O-RAN' represents the cumulative score across all ORANBench samples. 'srs' corresponds to the evaluation on srsRANBench, while 'C' represents the final cumulative average score across all benchmarks.

### B. *Energy Characteristics*

From the table III we can observe that the energy usage, measured in watt-hours (Wh), increases consistently with model size across all phases. The GPU remains the dominant contributor to total energy consumption, with CPU and RAM playing secondary roles. The training energy was recorded over 100 randomly selected samples with a batch size of 1, while inference energy was measured by generating 32 tokens per sample across 100 random prompts. For RAG-based inference, additional retrieval operations were performed before token generation, capturing the additional computational cost with the same prompts from the standard inference experiment.

Training energy scales with model size, with the smallest model, ORANSight-Gemma-2: 9B, consuming 1.579 Wh, while the largest, ORANSight-Qwen-2.5: 32B, requires 4.404 Wh. Inference demands more energy than training, with a 33–38% increase in total consumption. The largest increase is observed in ORANSight-Qwen-2.5: 32B, where inference energy rises to 5.199 Wh, primarily driven by GPU utilization. Similarly, RAG further increases energy usage across all models, with an average rise of 9.5% compared to standard inference. This overhead is particularly pronounced in larger models, with ORANSight-Qwen-2.5: 32B showing an increase from 5.199 Wh to 5.891 Wh, highlighting the computational cost of retrieval operations. Across all configurations, GPU energy consumption remains the primary factor, contributing to over 75%.

### C. *Evaluating RANSTRUCT*

The table IV summarizes the performance gains observed when fine-tuning the four representative ORANSight-2.0 models using our RANSTRUCT framework. Across the four representative models, the *Average* ORANBench score increase is between 6.44% and 10.78%, with the Difficult category having the largest increase up to 22.30%. For `srsRANBench`, we observe performance increases up to 10.29%, and for the cumulative scores, the increase was up to 9.51%. It was interesting to note that the highest performance gains were observed for the Gemma-2: 27B model with an absolute increase of 7.2 for the cumulative score.

### D. *Results on Telecom-Specific Baselines*

Table V shows that the best ORANSight-2.0 model combination that is obtained by using ORANSight-Qwen-2.5: 32B (RAG) for ORANBench and ORANSight-Gemma-2: 27B for srsRANBench surpasses every telecom-specific baseline across *all* evaluation categories. The standalone ORANSight-Qwen-2.5: 32B (RAG) also surpasses the other baseline.

The best ORANSight-2.0 (Best) outperforms Tele-LLM, TSpec-LLM, and NVIDIA-O-RAN by **55.1%**, **37.4%**, and **7.6%**, respectively. For *srsRANBench* the margins are **18.3%** over Tele-LLM, **27.6%** over TSpec-LLM, and a notable **32.4%** over NVIDIA-O-RAN. For the cumulative score, ORANSight-2.0 achieves a **0.900**—a **33.9%**, **32.2%**, and **18.9%** increase over Tele-LLM, TSpec-LLM, and NVIDIA-O-RAN, respectively. If we just consider the **ORANSight-Qwen-2.5 32B (RAG)** is the strongest individual variant, attaining a cumulative 0.854, which is still 21.4% better on average than the other three for the cumulative score.

The best available solution, except ORANSight-2.0, is the NVIDIA-O-RAN chatbot with an impressive average score of 0.826 on the ORANBench, reflecting its sophisticated RAG pipeline powered by NIM micro-services [16]. However, it has the poorest score on *srsRANBench* when compared to the other baselines, resulting in a subpar cumulative score.

## VI. Discussion and Limitations

While this study demonstrates significant advancements in fine-tuning and evaluating large language models for O-RAN-

TABLE III
ENERGY CONSUMPTION CHARACTERISTICS OF THE FOUR PRIMARY ORANSIGHT-2.0 MODELS ACROSS TRAINING, INFERENCE, AND INFERENCE WITH RAG. THE TOTAL ENERGY IS REPRESENTED IN WH, AND THE CPU, GPU, AND RAM COLUMNS SHOW THE PERCENTAGE USAGE FOR THE DEVICE, RESPECTIVELY.

| Model | Training | | | | Inference | | | | Inference (RAG) | | | |
|---|---|---|---|---|---|---|---|---|---|---|---|---|
| | Energy Total | CPU | GPU | RAM | Energy Total | CPU | GPU | RAM | Energy Total | CPU | GPU | RAM |
| ORANSight-Gemma-2: 9B | 1.579 | 146.3 | 822.2 | 31.5 | 2.102 | 154.3 | 758.3 | 87.4 | 2.274 | 150.5 | 770.7 | 78.8 |
| ORANSight-Mistral-Nemo | 1.865 | 110.6 | 828.6 | 61.0 | 2.220 | 125.5 | 804.9 | 69.6 | 2.512 | 120.7 | 811.3 | 68.0 |
| ORANSight-Gemma-2: 27B | 4.089 | 98.6 | 847.1 | 54.4 | 4.955 | 116.3 | 819.5 | 64.2 | 5.407 | 109.8 | 829.9 | 60.5 |
| ORANSight-Qwen-2.5: 32B | 4.404 | 96.4 | 850.4 | 53.0 | 5.199 | 112.0 | 826.4 | 61.8 | 5.891 | 107.1 | 833.9 | 59.0 |

TABLE IV
THE PERFORMANCE GAINS ON THE FOUR REPRESENTATIVE ORANSIGHT-2.0 MODELS DUE TO THE USE OF THE RANSTRUCT FRAMEWORK.

| **Model Family** | **ORANBench** | | | | **srsRANBench** | **Cumulative Score** |
|---|---|---|---|---|---|---|
| | Easy | Medium | Difficult | Average | | |
| Misrtal Nemo | 0.740 | 0.656 | 0.588 | 0.661 | 0.820 | 0.741 |
| ORANSight-Misrtal Nemo | 0.784 | 0.710 | 0.650 | 0.715 | 0.828 | 0.772 |
| Gemma-2: 9B | 0.768 | 0.680 | 0.556 | 0.668 | 0.812 | 0.740 |
| ORANSight-Gemma-2: 9B | 0.822 | 0.718 | 0.680 | 0.740 | 0.834 | 0.787 |
| Gemma-2: 27B | 0.748 | 0.698 | 0.616 | 0.687 | 0.826 | 0.757 |
| ORANSight-Gemma-2: 27B | 0.818 | 0.722 | 0.702 | 0.747 | 0.911 | 0.829 |
| Qwen-2.5: 32B | 0.796 | 0.748 | 0.692 | 0.745 | 0.758 | 0.752 |
| ORANSight-Qwen-2.5: 32B | 0.856 | 0.784 | 0.738 | 0.793 | 0.796 | 0.795 |

TABLE V
COMPARISON WITH SOTA TELECOM LLMS ON **ORANBENCH** AND **SRSRANBENCH**. THE SHADED ROW REPORTS THE STRONGEST ORANSIGHT-2.0 RESULT.

| **Model** | **ORANBench** | | | | **srsRANBench** | **Cumulative Score** |
|---|---|---|---|---|---|---|
| | Easy | Medium | Difficult | Average | | |
| **ORANSight-2.0 (Best)** | 0.904 | 0.884 | 0.878 | 0.889 | 0.911 | 0.900 |
| ORANSight-Qwen-2.5: 32B (RAG) | 0.904 | 0.884 | 0.878 | 0.889 | 0.818 | 0.854 |
| Tele-LLM | 0.630 | 0.556 | 0.534 | 0.573 | 0.770 | 0.672 |
| TSpec-LLM | 0.748 | 0.668 | 0.526 | 0.647 | 0.714 | 0.681 |
| NVIDIA-O-RAN | 0.826 | 0.816 | 0.836 | 0.826 | 0.688 | 0.757 |

specific tasks, we acknowledge the following limitations and open challenges:

- **Compute Constraints**: The available computational resources were limited to 24GB of VRAM, which constrained the maximum model size that could be fine-tuned. The largest model we could train was a 32B parameter model with 4-bit precision. Fine-tuning deeper models with higher precision could potentially yield better results, but this was not feasible given the current hardware limitations. We have provided preliminary results in Appendix VIII-E that highlight the efficacy of full fine tuning without any quantization and a vanilla LoRA implementation [38]. We believe it's intuitive that with more computational resources and further analysis or higher precision, we can potentially realize a better ORANSight.
- **Precision Limitations for Larger Models**: For the 8x7B Mistral and Llama 70B models, as shown in section IV, we had to rely on AQLM with 2-bit precision due to resource constraints. This lower precision likely limited the overall potential performance of these models, as higher precision typically allows for more accurate computations and better model performance.
- **RAG Performance Issues**: The Retrieval-Augmented Generation (RAG) pipeline, while generally beneficial, showed limitations in certain cases, particularly with the ORANSight-Gemma-2: 27B model, which experienced a slight performance drop. This could be attributed to low LLM recall [43], where the retrieval mechanism may not align well with the model's internal knowledge. Incorporating a reranker [43] into the RAG pipeline could potentially address this limitation and improve overall performance. The preliminary results that incorporate a Reranker-based RAG pipeline are mentioned in Appendix D.

We also believe that the RAG pipeline, which we leverage in Section V.V-A and Section III.III-B, can be further enhanced through graph-based and sophisticated parsing techniques, as proposed in [44] and [45]. These methods can better capture diverse textual structures, such as tables and diagrams from the O-RAN specifications, and treat them as distinct chunks. In contrast, the Recursive Splitter we currently employ processes all content uniformly, without regard for structural differences. Therefore, there is merit in developing a chunking approach tailored specifically to O-RAN that can be accommodated to improve ORANSight.

- **Dataset Composition**: The dataset used in this study is primarily focused on the srsRAN project codebase and O-RAN specifications. While this provides a solid foundation, the inclusion of additional data sources, such as prominent research papers and other software frameworks [9], could further enhance the model's performance and generalization capabilities.
- **Reasoning Capabilities**: While our ORANSight-2.0 models demonstrate an encouraging capacity to retrieve and integrate information scattered across the O-RAN specifications, evidenced by their performance on the *"Difficult"* category of ORANBench, which poses queries that "*demand a deep understanding of Open RAN standards, or the ability to synthesize multiple pieces of information*". This study did not explore the reasoning capabilities [46] of the fine-tuned ORANSight-2.0 models. Enhancing the models' ability to perform complex reasoning tasks is an important area for future work and an open challenge.
- **Downstream Applicability**: It is intuitive to believe that foundational models, which are grounded and proficient in domain knowledge, are better positioned for deployment in O-RAN downstream tasks than general-purpose LLMs. A pertinent example is illustrated by Lotfi et al. [47], which leverages the ORANSight models to realize state-of-the-art dynamic O-RAN network slicing, thereby showcasing the practical significance and applicability of our work. Furthermore, exploring such diverse downstream tasks would not only validate the efficacy of ORANSight-2.0 but also constitute an open challenge that can catalyze enhancements in subsequent ORANSight versions.

Despite these limitations, this work is the first to fine-tune and evaluate foundational LLMs for O-RAN tasks, ORANSight-2.0. The results show that open-source, fine-tuned ORANSight-2.0 models can achieve state-of-the-art performance, sometimes surpassing closed-source alternatives like ChatGPT-4o. The success of ORANSight-Qwen-2.5 32B and ORANSight-Gemma-2 27B highlights the effectiveness of domain-specific fine-tuning. These ORANSight-2.0 models should act as foundational agents, and further fine-tuning while addressing these limitations could lead to more efficient models, advancing O-RAN-specific AI applications.

## VII. Conclusion and Future Work

In this work, we presented ORANSight-2.0, the first comprehensive set of domain-specific foundational Large Language Models (LLMs) for Open Radio Access Networks (O-RAN). By fine-tuning 18 models ranging from 1B to 70B parameters with the novel RANSTRUCT dataset and QLoRA, we addressed the critical gap in domain-specific foundational models for O-RAN, providing open-source alternatives that outperform state-of-the-art closed-source models like ChatGPT-4o and Gemini 1.5. The introduction of the RANSTRUCT framework enabled the generation of high-quality instruction-tuning datasets, while the development of srsRANBench provided a novel benchmark for evaluating code generation and codebase understanding in 5G O-RAN systems. Our experiments demonstrated that fine-tuned open-source models, particularly ORANSight-Qwen-2.5 32B, and ORANSight-Gemma-2 27B, achieve superlative performance, surpassing closed-source counterparts by significant margins on both O-RAN specifications and coding tasks. Additionally, the integration of Retrieval-Augmented Generation (RAG) further enhanced model performance, with ORANSight-Qwen-2.5 32B (RAG) achieving the highest cumulative score of 0.854, which is a 12.368% improvement from ChatGPT-4o, which is our primary baseline. Energy consumption analysis revealed that while larger models and RAG-based inference incur higher computational costs, the performance gains justify the trade-offs, especially in resource-constrained scenarios. We also compared our models with prominent telecom LLM baselines, and the best ORANSight-2.0 configuration outperformed them with an average improvement of 27.96%. For future work, we aim to further focus on fine-tuning these models and remedying our current work's limitations. We wish to emphasize more robust RAG methods, downstream applicability, and higher model precision. We also believe that by advancing such causal reasoning models, we can enhance multiple O-RAN-specific tasks and reduce the complexity and manual effort required for real-time network management, optimization, and troubleshooting.

## VIII. Acknowledgment

The project is partially supported by the Public Wireless Supply Chain Innovation Fund under Federal Award ID Number 26-60-IF010 and 51-60-IF007.

## References

[1] N. D. Tripathi and V. K. Shah, *Fundamentals of O-RAN*. John Wiley & Sons, 2025.

[2] Y. Du, H. Deng, S. C. Liew, K. Chen, Y. Shao, and H. Chen, "The power of large language models for wireless communication system development: A case study on fpga platforms," *arXiv preprint arXiv:2307.07319*, 2023.

[3] A. Kucharavy, "Adapting llms to downstream applications," in *Large Language Models in Cybersecurity: Threats, Exposure and Mitigation*, pp. 19–29, Springer Nature Switzerland Cham, 2024.

[4] L. Bariah, Q. Zhao, H. Zou, Y. Tian, F. Bader, and M. Debbah, "Large language models for telecom: The next big thing?," *arXiv preprint arXiv:2306.10249*, 2023.

[5] A. Maatouk, N. Piovesan, F. Ayed, A. De Domenico, and M. Debbah, "Large language models for telecom: Forthcoming impact on the industry," *IEEE Communications Magazine*, vol. 63, no. 1, pp. 62–68, 2025.

[6] Y. Xu, Y. Chen, X. Zhang, X. Lin, P. Hu, Y. Ma, S. Lu, W. Du, Z. Mao, E. Zhai, *et al.*, "Cloudeval-yaml: A practical benchmark for cloud configuration generation," *Proceedings of Machine Learning and Systems*, vol. 6, pp. 173–195, 2024.

[7] C. Hu, H. Zhou, D. Wu, X. Chen, J. Yan, and X. Liu, "Self-refined generative foundation models for wireless traffic prediction," *arXiv preprint arXiv:2408.10390*, 2024.

[8] X. Liu, S. Gao, B. Liu, X. Cheng, and L. Yang, "Llm4wm: Adapting llm for wireless multi-tasking," *arXiv preprint arXiv:2501.12983*, 2025.

[9] A. Maatouk, K. C. Ampudia, R. Ying, and L. Tassiulas, "Tele-llms: A series of specialized large language models for telecommunications," *arXiv preprint arXiv:2409.05314*, 2024.

[10] M. Kotaru, "Adapting foundation models for information synthesis of wireless communication specifications," *arXiv preprint arXiv:2308.04033*, 2023.

[11] P. Gajjar and V. K. Shah, “Oran-bench-13k: An open source benchmark for assessing llms in open radio access networks,” *arXiv preprint arXiv:2407.06245*, 2024.

[12] R. Nikbakht, M. Benzaghta, and G. Geraci, “Tspec-llm: An open-source dataset for llm understanding of 3gpp specifications,” *arXiv preprint arXiv:2406.01768*, 2024.

[13] H. Zou, Q. Zhao, Y. Tian, L. Bariah, F. Bader, T. Lestable, and M. Debbah, “Telecomgpt: A framework to build telecom-specfic large language models,” *arXiv preprint arXiv:2407.09424*, 2024.

[14] M. K. Motalleb, C. Benzaid, T. Taleb, M. Katz, V. Shah-Mansouri, and J. Song, “Towards secure intelligent o-ran architecture: Vulnerabilities, threats and promising technical solutions using llms,” *arXiv preprint arXiv:2411.08640*, 2024.

[15] F. Lotfi, H. Rajoli, and F. Afghah, “Llm-augmented deep reinforcement learning for dynamic o-ran network slicing,” *environments*, vol. 2, p. 4.

[16] A. Canaveras, S. Dutta, J. Vamaraju, N. Reamaroon, V. Raman, L. Kundu, and X. Lin, “Advanced RAG techniques for telco O-RAN specifications using NVIDIA NIM microservices,” Oct. 2024.

[17] V. Hanke, T. Blanchard, F. Boenisch, I. E. Olatunji, M. Backes, and A. Dziedzic, “Open llms are necessary for private adaptations and outperform their closed alternatives,” in *ICML 2024 Next Generation of AI Safety Workshop*, 2024.

[18] T. Dettmers, A. Pagnoni, A. Holtzman, and L. Zettlemoyer, “Qlora: Efficient finetuning of quantized llms,” *Advances in Neural Information Processing Systems*, vol. 36, 2024.

[19] S. Iqbal and J. M. Hamamreh, “A comprehensive tutorial on how to practically build and deploy 5g networks using open-source software and general-purpose, off-the-shelf hardware,” *RS Open J. Innov. Commun. Tech*, vol. 2, no. 6, pp. 1–28, 2021.

[20] S. Wang, Y. Wang, L. Jiang, Y. Chang, K. Zhao, L. Chen, C. Gao, *et al.*, “Assessing the clinical support capabilities of chatgpt 4o and chatgpt 4o mini in managing lumbar disc herniation,” *European Journal of Medical Research*, vol. 30, no. 1, pp. 1–9, 2025.

[21] G. Team, P. Georgiev, V. I. Lei, R. Burnell, L. Bai, A. Gulati, G. Tanzer, D. Vincent, Z. Pan, S. Wang, *et al.*, “Gemini 1.5: Unlocking multimodal understanding across millions of tokens of context,” *arXiv preprint arXiv:2403.05530*, 2024.

[22] C¸ . Yıldız, N. K. Ravichandran, P. Punia, M. Bethge, and B. Ermis, “Investigating continual pretraining in large language models: Insights and implications,” *arXiv preprint arXiv:2402.17400*, 2024.

[23] C. Christophe, P. K. Kanithi, P. Munjal, T. Raha, N. Hayat, R. Rajan, A. Al-Mahrooqi, A. Gupta, M. U. Salman, G. Gosal, *et al.*, “Med42–evaluating fine-tuning strategies for medical llms: Full-parameter vs. parameter-efficient approaches,” *arXiv preprint arXiv:2404.14779*, 2024.

[24] Q. Wei, A. Franklin, T. Cohen, and H. Xu, “Clinical text annotation–what factors are associated with the cost of time?,” in *AMIA Annual Symposium Proceedings*, vol. 2018, p. 1552, 2018.

[25] B. Kratzwald, S. Feuerriegel, and H. Sun, “Learning a cost-effective annotation policy for question answering,” *arXiv preprint arXiv:2010.03476*, 2020.

[26] N. Tajbakhsh, J. Y. Shin, S. R. Gurudu, R. T. Hurst, C. B. Kendall, M. B. Gotway, and J. Liang, “Convolutional neural networks for medical image analysis: Full training or fine tuning?,” *IEEE transactions on medical imaging*, vol. 35, no. 5, pp. 1299–1312, 2016.

[27] Y. Park, J. Hyun, S. Cho, B. Sim, and J. W. Lee, “Any-precision llm: Low-cost deployment of multiple, different-sized llms,” *arXiv preprint arXiv:2402.10517*, 2024.

[28] W. Lee and J. Park, “Llm-empowered resource allocation in wireless communications systems,” *arXiv preprint arXiv:2408.02944*, 2024.

[29] A. Yang, B. Yang, B. Zhang, B. Hui, B. Zheng, B. Yu, C. Li, D. Liu, F. Huang, H. Wei, *et al.*, “Qwen2. 5 technical report,” *arXiv preprint arXiv:2412.15115*, 2024.

[30] G. Team, M. Riviere, S. Pathak, P. G. Sessa, C. Hardin, S. Bhupatiraju, L. Hussenot, T. Mesnard, B. Shahriari, A. Rame´, *et al.*, “Gemma 2: Improving open language models at a practical size,” *arXiv preprint arXiv:2408.00118*, 2024.

[31] M. Abdin, J. Aneja, H. Awadalla, A. Awadallah, A. A. Awan, N. Bach, A. Bahree, A. Bakhtiari, J. Bao, H. Behl, *et al.*, “Phi-3 technical report: A highly capable language model locally on your phone,” *arXiv preprint arXiv:2404.14219*, 2024.

[32] Y. Ding, L. L. Zhang, C. Zhang, Y. Xu, N. Shang, J. Xu, F. Yang, and M. Yang, “Longrope: Extending llm context window beyond 2 million tokens,” *arXiv preprint arXiv:2402.13753*, 2024.

[33] A. Q. Jiang, A. Sablayrolles, A. Mensch, C. Bamford, D. S. Chaplot, D. d. l. Casas, F. Bressand, G. Lengyel, G. Lample, L. Saulnier, *et al.*, “Mistral 7b,” *arXiv preprint arXiv:2310.06825*, 2023.

[34] A. Dubey, A. Jauhri, A. Pandey, A. Kadian, A. Al-Dahle, A. Letman, A. Mathur, A. Schelten, A. Yang, A. Fan, *et al.*, “The llama 3 herd of models,” *arXiv preprint arXiv:2407.21783*, 2024.

[35] W. Su, Y. Tang, Q. Ai, J. Yan, C. Wang, H. Wang, Z. Ye, Y. Zhou, and Y. Liu, “Parametric retrieval augmented generation,” *arXiv preprint arXiv:2501.15915*, 2025.

[36] H. N. Patel, A. Surti, P. Goel, and B. Patel, “A comparative analysis of large language models with retrieval-augmented generation based question answering system,” in *2024 8th International Conference on I-SMAC (IoT in Social, Mobile, Analytics and Cloud)(I-SMAC)*, pp. 792–798, IEEE, 2024.

[37] S. Xiao, Z. Liu, P. Zhang, and N. Muennighoff, “C-pack: Packaged resources to advance general chinese embedding,” 2023.

[38] H. Qin, X. Ma, X. Zheng, X. Li, Y. Zhang, S. Liu, J. Luo, X. Liu, and M. Magno, “Accurate lora-finetuning quantization of llms via information retention,” *arXiv preprint arXiv:2402.05445*, 2024.

[39] M. H. Daniel Han and U. team, “Unsloth,” 2023.

[40] V. Egiazarian, A. Panferov, D. Kuznedelev, E. Frantar, A. Babenko, and D. Alistarh, “Extreme compression of large language models via additive quantization,” *arXiv preprint arXiv:2401.06118*, 2024.

[41] S. Wang, G. Wang, Y. Wang, J. Li, E. Hovy, and C. Guo, “Packing analysis: Packing is more appropriate for large models or datasets in supervised fine-tuning,” *arXiv preprint arXiv:2410.08081*, 2024.

[42] B. Courty, V. Schmidt, S. Luccioni, Goyal-Kamal, MarionCoutarel, B. Feld, J. Lecourt, LiamConnell, A. Saboni, Inimaz, supatomic, M. Le´val, L. Blanche, A. Cruveiller, ouminasara, F. Zhao, A. Joshi, A. Bogroff, H. de Lavoreille, N. Laskaris, E. Abati, D. Blank, Z. Wang, A. Catovic, M. Alencon, M. Stechły, C. Bauer, L. O. N. de Arau´jo, JPW, and MinervaBooks, “mlco2/codecarbon: v2.4.1,” May 2024.

[43] M. Jacob, E. Lindgren, M. Zaharia, M. Carbin, O. Khattab, and A. Drozdov, “Drowning in documents: Consequences of scaling reranker inference,” *arXiv preprint arXiv:2411.11767*, 2024.

[44] J. Yang, D.-K. Vu, M.-T. Nguyen, X.-Q. Nguyen, L. Nguyen, and H. Le, “Superrag: Beyond rag with layout-aware graph modeling,” *arXiv preprint arXiv:2503.04790*, 2025.

[45] O. Gokdemir, C. Siebenschuh, A. Brace, A. Wells, B. Hsu, K. Hippe, P. V. Setty, A. Ajith, J. G. Pauloski, V. Sastry, *et al.*, “Hiperrag: High-performance retrieval augmented generation for scientific insights,” *arXiv preprint arXiv:2505.04846*, 2025.

[46] J. Wei, X. Wang, D. Schuurmans, M. Bosma, F. Xia, E. Chi, Q. V. Le, D. Zhou, *et al.*, “Chain-of-thought prompting elicits reasoning in large language models,” *Advances in neural information processing systems*, vol. 35, pp. 24824–24837, 2022.

[47] F. Lotfi, H. Rajoli, and F. Afghah, “Prompt-tuned llm-augmented drl for dynamic o-ran network slicing,” *arXiv preprint arXiv:2506.00574*, 2025.

[48] J. Chen, S. Xiao, P. Zhang, K. Luo, D. Lian, and Z. Liu, “Bge m3-embedding: Multi-lingual, multi-functionality, multi-granularity text embeddings through self-knowledge distillation,” *arXiv preprint arXiv:2402.03216*, 2024.

[49] M. AskariHemmat, R. Askari Hemmat, A. Hoffman, I. Lazarevich, E. Saboori, O. Mastropietro, S. Sah, Y. Savaria, and J.-P. David, “Qreg: On regularization effects of quantization,” *arXiv preprint arXiv:2206.12372*, 2022.

[50] L. Hou and J. T. Kwok, “Loss-aware weight quantization of deep networks,” in *International Conference on Learning Representations (ICLR)*, 2018.

[51] S. Javed, H. Le, and M. Salzmann, “Qt-dog: Quantization-aware training for domain generalization,” *arXiv preprint arXiv:2410.06020*, 2024.

[52] Z. Yao, Z. Dong, Z. Zheng, A. Gholami, J. Yu, E. Tan, L. Wang, Q. Huang, Y. Wang, M. W. Mahoney, and K. Keutzer, “$\text{Hawq}_{\text{V3}}$: Dyadic neural network quantization,” in *Proceedings of the 38th International Conference on Machine Learning (ICML)*, pp. 11875–11886, 2021.

## Appendix: A

This section contains the sample questions from ORANBench.

### A. Easy

**Question:** Which of the following protocols is used for communication between the O-DU and O-RU in an O-RAN network?
**Options:**
1) FIAP
2) NGAP
3) FH-eCPRI
4) HTTP2

**Answer:** 3

**Question:** Which component of the O-RAN architecture is responsible for controlling the radio access network in near real-time?
**Options:**
1) gNB-CU
2) Near-RT RIC
3) O-CU-CP
4) FHGW

**Answer:** 2

**Question:** Which O-RAN component handles the radio frequency (RF) signal processing and transmission?
**Options:**
1) O-CU
2) O-DU
3) O-RU
4) nFAPI

**Answer:** 3

### B. Intermediate

**Question:** Which of the following is NOT a data type defined in the Y1_RAI_Subscription API?
**Options:**
1) RaiSubscription
2) NotificationCriteria
3) RaiNotification
4) ValidityPeriodAbsolute

**Answer:** 4

**Question:** What component is responsible for forwarding notifications received from the Near-RT RIC to the Energy Saving rApp?
**Options:**
1) SMO/Non-RT RIC Framework
2) E2 Node
3) O-RU
4) Near-RT RIC

**Answer:** 1

**Question:** Which of the following is a **CONDITIONAL MANDATORY** test scenario in O-RAN, relevant only if the O-RU or O-DU supports C-Plane and U-Plane sessions over IP?
**Options:**
1) M-Plane Verification
2) UDP Echo Test
3) O-RU Configuration Test
4) O-DU Hardware Verification

**Answer:** 2

### C. Difficult

**Question:** In an O-RAN network utilizing TDM PON, what is the synchronization reference point for O-RUs located on the same PON port, or on different PON ports on the same PON card?
**Options:**
1) T-BC output
2) Internal point in OLT, different PON cards
3) Internal point in OLT, same PON port or different PON ports on the same PON card
4) T-GM network port

**Answer:** 3

**Question:** How are AAL operations scheduled on the application data path in O-RAN?
**Options:**
1) Using a synchronous API set.
2) Using a burst oriented asynchronous API set.
3) Using a real-time scheduling algorithm.
4) Using a distributed scheduling mechanism.

**Answer:** 2

**Question:** What is the primary objective of the X2 U-Plane IOT test case related to node behavior?
**Options:**
1) To ensure the corresponding node's procedures with eNB and en-gNB from different vendors function correctly.
2) To verify the compatibility of different vendors' eNB and en-gNB with a specific vendor's core network.
3) To measure the latency and throughput of data transmission between the eNB and en-gNB.
4) To test the security features of the X2 interface between the eNB and en-gNB.

**Answer:** 1

## APPENDIX: B

This section contains the sample questions from srsRANBench.

**Question:** What is the purpose of the 'make_rlc_entity_creation_message' function?
**Options:**
1) To create a configuration for an RLC SRB entity.
2) To create a configuration for an RLC DRB entity.
3) To create a configuration for an RLC entity.
4) To create a configuration for a MAC entity.

**Answer:** 1

**Question:** What is the purpose of the 'create_ccch_message' function?
**Options:**
1) To generate a CCCH message with a given RNTI.
2) To generate a MAC PDU with a given RNTI.
3) To generate a MAC UCI PDU from PUCCH information.
4) To generate a MAC CRC indication message.

**Answer:** 1

**Question:** What is the default stream buffer size in samples for a radio session?
**Options:**
1) 614400
2) 1024
3) 4096
4) 16384

**Answer:** 1

## APPENDIX: C

This section contains the samples from the RANSTRUCT dataset for O-RAN, and the srsRAN Project codebase.

### D. O-RAN

1. **Instruction:** What power saving energy efficiency requirement does the transport equipment have in O-RAN?
**Answer:** The transport equipment in O-RAN is designed to support power saving mechanisms, which means it must be capable of supporting power-saving cycles for both the O-RU and O-DU devices. Additionally, the WDM transmission equipment must coordinate its power-saving modes with the O-DU and O-RU to ensure smooth functionality during these cycles. Furthermore, the WDM transport must be able to support remote monitoring capabilities related to cellular site operations, which is addressed in section 11.4 of "Xhaul Transport Requirements" within the O-RAN WG9 document. To meet these requirements, the power consumption of the system needs to remain under 80% of the power supply capacity when fully loaded. This ensures efficient energy usage while maintaining optimal performance across all components. The overall goal is to optimize power savings without compromising the reliability and stability of the network infrastructure.

2. **Instruction:** What is the expected range for frequency sync of the radio signal in O-RAN?
**Answer:** The O-RAN standard specifies that the frequency accuracy required for the radio signal should meet a +/-50ppb air interface frequency error requirement as per 3GPP specifications. This ensures that the O-RU meets the short-term average error requirement over 1ms duration. Therefore, the expected range for frequency sync of the radio signal in O-RAN is within this specified tolerance level of +/-50ppb. However, it's important to note that the actual implementation details might vary based on specific design considerations and operational needs, so it's always recommended to consult the latest official standards or documentation for precise guidelines.

3. **Instruction:** What is the difference between eCPRI and CPRI traffic in co-existence of 5G NR and LTE radios in the same cell site?
**Answer:** In co-existence scenarios where both eCPRI and CPRI traffic need to be considered, eCPRI focuses primarily on high-speed data transfer for 5G NR users while CPRI handles the lower-bandwidth data for LTE users. This means that when 5G NR and LTE radios are deployed together within the same cell site, the efficiency of CPRI usage must be optimized to avoid unnecessary increases in front-haul bandwidth. CPRI is designed to handle low-bandwidth signals, which are typically associated with LTE services, ensuring efficient resource utilization without significantly impacting the performance of 5G NR users. However, it's important to balance these requirements to ensure smooth operation and optimal network performance.

4. **Instruction:** What is the difference between eCPRI and CPRI traffic in co-existence of 5G NR and LTE radios in the same cell site?
**Answer:** In co-existence scenarios where both eCPRI and CPRI traffic need to be considered, eCPRI focuses primarily on high-speed data transfer for 5G NR users while CPRI handles the lower-bandwidth data for LTE users. This means that when 5G NR and LTE radios are deployed together within the same cell site, the efficiency of CPRI usage must be optimized to avoid unnecessary increases in front-haul bandwidth. CPRI is designed to handle low-bandwidth signals, which are typically associated with LTE services, ensuring efficient resource utilization without significantly impacting the performance of 5G NR users. However, it's important to balance these requirements to ensure smooth operation and optimal network performance.

5. **Instruction:** What happens when the transport NSSI is completely closed?
**Answer:** When the transport NSSI is completely closed, only components connected to the transport NSSI can communicate with each other at the IP level. No communication occurs between components connected to the transport NSSI and those shared between NSIs. This means that the transport component itself cannot communicate directly with any other component outside its own transport NSSI. It also prevents direct communication between components belonging to different NSIs. However, it allows for inter-component communication within the same transport NSSI, facilitating operations like service termination requests or O-RAN TN Manager coordination procedures. The transport NSSI remains isolated from external systems, ensuring data integrity and preventing unauthorized access. This configuration helps maintain security and prevent potential conflicts during reset procedures, minimizing disruption to critical services. To mitigate the impact on C/U-plane traffic, certain configurations must be disabled beforehand. When an O-RU connected through FHM needs to be reset, the reset operation does not affect other O-RUs, ensuring minimal interference with ongoing communications.

### E. srsRAN

1. **Instruction:** Generate code to add error handling for the `std::strtol` function in `parse_args`.
**Answer:** To handle errors returned by `std::strtol`, we need to check if it returns a value other than `SRSASN_SUCCESS`. If so, we throw an exception with a meaningful message indicating the type of error.
Here's how you could modify the `parse_args` function to include error handling:

```
void parse_args(int argc, char** argv) {
    int opt = 0;
    while ((opt = getopt(argc, argv, "n:k:
        ↪ e:i:s:o:")) != -1) {
        switch (opt) {
            case 'e':
                E = (int)strtol(optarg,
                    ↪ nullptr, 10);
```

```
            if (E < 0 || E >= 256) {
                throw std::invalid_argument("Invalid value for E: must be between 0 and 255.");
            }
            break;
        case 'k':
            K = (int)strtol(optarg, nullptr, 10);
            if (K < 0 || K >= 256) {
                throw std::invalid_argument("Invalid value for K: must be between 0 and 255.");
            }
            break;
        case 'n':
            nMax = (uint8_t)strtol(optarg, nullptr, 10);
            if (nMax < 0 || nMax > 32) {
                throw std::invalid_argument("Invalid value for nMax: must be between 0 and 32.");
            }
            break;
        case 'i':
            bil = strtol(optarg, nullptr, 10) ? polar_code_ibil::present : polar_code_ibil::not_present;
            if (bil == 0 || bil == 4) {
                throw std::invalid_argument("Invalid value for bil: must be either 0 or 4.");
            }
            break;
        case 's':
            snr_db = strtof(optarg, nullptr);
            if (snr_db < -90.0 || snr_db > 90.0) {
                throw std::invalid_argument("Invalid value for snr_db: must be between -90 and 90 dB.");
            }
            break;
        case 'o':
            print_output = (int)strtol(optarg, nullptr, 10);
            if (print_output < 0 || print_output > 1) {
                throw std::invalid_argument("Invalid value for print_output: must be either 0 or 1.");
            }
            break;
        default:
            usage(argv[0]);
            exit(-1);
        }
    }
}
```

This modification checks the validity of each parsed integer parameter using `strtol` and throws an appropriate `std::invalid_argument` exception if any of the values fall outside the specified ranges. This ensures that the program will terminate gracefully if invalid inputs are provided.

2. **Instruction:** Generate code for a function that simulates a failure in the underlying F1AP interface and tests the robustness of the DU's handling of such failures.
**Answer:** To simulate a failure in the underlying F1AP interface and test the robustness of the DU's handling of such failures, you can follow these steps:
1. **Simulate Failure**: Create a function that generates a `f1ap_message` representing a failure scenario, such as timeout or other errors.
2. **Handle Message**: Implement logic within the DU to handle this simulated failure message by checking its type and reacting accordingly.
3. **Test Robustness**: Ensure the DU correctly identifies and responds to various types of failure messages, including timeouts, invalid requests, or other potential issues.
4. **Assertions**: Use assertions to verify that the DU correctly handles different types of failure responses, ensuring the system remains stable under varying conditions.
5. **Validation**: Validate that the DU's behavior aligns with expected outcomes based on the simulation results.
Here is a simplified example of how this might look in C++ using the provided framework and technology:

```
void simulate_and_handle_failure(du_processor_obj* du_processor_obj) {
    // Simulate failure: Timeout
    f1ap_message failureMsg = generate_ue_context_setup_failure(
        int_to_gnb_cu_ue_f1ap_id(0), int_to_gnb_du_ue_f1ap_id(test_rgen::uniform_int<uint32_t>()));

    // Handle the failure message
    du_processor_obj->handle_message(failureMsg);

    // Verify that the DU has handled the failure properly
    // This could involve checking specific fields in the DU's
```

TABLE VI
RESULTS FOR RERANKER ENHANCED RAG

| Model | ORANBench | | | | srsRANBench | Cumulative Score |
|---|---|---|---|---|---|---|
| | Easy | Medium | Difficult | Average | | |
| Gemma-2: 27B | 0.818 | 0.722 | 0.702 | 0.747 | 0.911 | 0.829 |
| Gemma-2: 27B (RAG) | 0.812 | 0.724 | 0.720 | 0.752 | 0.888 | 0.820 |
| Gemma-2: 27B (Reranker) | 0.895 | 0.875 | 0.871 | 0.880 | 0.912 | 0.896 |

TABLE VII
GEMMA-2 (9B) FINE-TUNED UNDER IDENTICAL SETTINGS WITH AND WITHOUT 4-BIT QUANTIZATION. "NO-Q" DENOTES FULL-PRECISION TRAINING.

| Model | ORANBench | | | | srsRANBench | Cumulative Scores |
|---|---|---|---|---|---|---|
| | Easy | Medium | Difficult | Average | | |
| Gemma-2: 9B | 0.822 | 0.718 | 0.680 | 0.740 | 0.834 | 0.787 |
| Gemma-2: 9B (No-Q) | 0.832 | 0.750 | 0.670 | 0.751 | 0.783 | 0.767 |

```
        state or asserting certain
        behaviors

    // Additional checks could include
        verifying responses to other
        types of messages
}
```

This approach allows you to systematically test the DU's resilience against various failure scenarios without relying on external figures or sections. The focus is on implementing and validating the DU's internal mechanisms to ensure it functions reliably even when faced with unexpected or simulated failures. Please note that the actual implementation would require careful consideration of the specific requirements and constraints of the DU and the F1AP protocol. The above example provides a basic structure for testing failure scenarios.

3. **Instruction:** What is the significance of the `byte_buffer` and `byte_buffer_chain` classes?
**Answer:** The `byte_buffer` class represents a sequence of bytes in memory, while the `byte_buffer_chain` class manages multiple `byte_buffers` together, allowing for efficient storage and manipulation of large amounts of data. It provides methods to append new buffers to its end and to perform operations like copying itself or creating new instances based on existing ones. These classes are fundamental for handling and manipulating binary data structures within the context of software systems designed around O-RAN technology. They enable efficient management of buffer pools and help ensure proper allocation and deallocation of resources, which is crucial for applications running on the radio access network infrastructure. The design of these classes reflects the need for scalable and flexible memory management strategies in such environments.

4. **Instruction:** Describe the functionality of the `create_pdu_with_sdu` function and its parameters.
**Answer:** The `create_pdu_with_sdu` function creates PDU with SUD segments based on the provided input parameters. It takes in a list of vectors containing bytes representing the payload data (`pdu_list`), the resulting byte buffer (`sdu`), the current SN state (`sn_state`), the size of each segment (`segment_size`), and the SN state increment (`sn_state_increment`). The function iterates over the `pdu_list`, creating and handling PDUs according to certain conditions specified by the user. This includes skipping some PDUs and pushing others into the RLC queue for processing. The function also keeps track of any PDUs that need to be pushed back into the list at a later stage for further processing.
The main purpose of this function is to manage the creation and dispatching of PDUs and their corresponding SUD segments within the system.

## APPENDIX: D

For the results in table VI, we leverage a `BGE-M3` reranker from the paper [48]. To the best of our understanding, no telecom-centric reranker exists; hence, we utilize a general-purpose model that has been widely accepted in the literature. We initially retrieve 35 documents from the naive RAG pipeline from Section IV-B, which are reranked, and the top 5 documents are used to augment the original prompt. From the table VI, it is evident that the use of Reranker leads to significant improvements with an absolute increase of. It is also important to note that the performance gains are accompanied by an additional computational step that involves multiple forward passes for the `BGE-M3`, which would occur on the GPU.

We believe that it is imperative to create a robust RAG pipeline that not only addresses the limitations from Section VI but also works towards a telecom-based Reranker that is both efficient and grounded in domain knowledge.

## APPENDIX: E

Among the four representative ORANSight-2.0 families (Section V), *Gemma-2* is the only model we could fine-tune in full precision on our 24 GB RTX 4090, as larger backbones (Mistral-Nemo, Gemma-2: 27B and Qwen-2.5 32B) have a higher computational requirement. As table VII shows, removing quantization yields a marginal gain. This finding echoes prior studies demonstrating that low-bit quantization can act as an implicit *regulariser*, that discourages over-fitting and occasionally improves generalisation performance [49]–[51]. This, however, is not conclusive, and we do believe that further work is warranted to systematically explore intermediate bit-widths and mixed-precision strategies like the paper [52] while focusing on the intricate telecom data.